\documentclass[runningheads]{llncs}
\usepackage[,mobile]{eccv}
\usepackage{eccvabbrv}
\usepackage{bbm}
\usepackage{graphicx}
\usepackage[normalem]{ulem}
\usepackage{subcaption}
\usepackage{float}
\usepackage{svg}
\usepackage{multicol}
\usepackage{multirow}
\usepackage{booktabs}
\usepackage[accsupp]{axessibility}
\usepackage[pagebackref,breaklinks,colorlinks,citecolor=eccvblue]{hyperref}
\usepackage{orcidlink}
\usepackage{comment}
\newcommand\ExtraSep
{\dimexpr\cmidrulewidth+\aboverulesep+\belowrulesep\relax}
\DeclareMathOperator*{\argmax}{arg\,max}

\begin{document}
\title{Learn ``No'' to Say ``Yes'' Better: Improving Vision-Language Models via Negations} 
\author{Jaisidh Singh\inst{1}\thanks{Denotes equal contribution.} \and
Ishaan Shrivastava\inst{1}{$^*$} \and
Mayank Vatsa\inst{1} \and Richa Singh\inst{1} \and Aparna Bharati\inst{2}}
\authorrunning{J.~Singh \and I.~Shrivastava et al.}
\institute{IIT Jodhpur, India \and
Lehigh University, USA\\
\email{\{singh.118, shrivastava.9, mvatsa, richa\}@iitj.ac.in}\\
\email{apb220@lehigh.edu}}
\maketitle
\vspace{-20pt}
\begin{figure*}
    \centering
    \includegraphics[width=\linewidth]{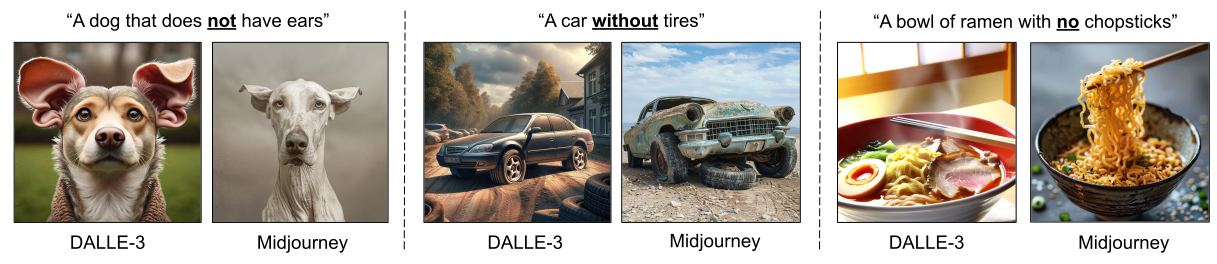}
    \caption{Vision-language models (VLMs) struggle to understand negations in text, observable in image-text matching and applications such as text-to-image generation~\cite{dalle3, midjourney}.\vspace{-30pt}}
    \label{fig:teaser}
\end{figure*}
\begin{abstract}
Existing vision-language models (VLMs) treat text descriptions as a unit, confusing individual concepts in a prompt and impairing visual semantic matching and reasoning.
An important aspect of reasoning in logic and language is negations.
This paper highlights the limitations of popular VLMs such as CLIP, at understanding the implications of negations, i.e., the effect of the word ``not'' in a given prompt. 
To enable evaluation of VLMs on fluent prompts with negations, we present CC-Neg, a dataset containing $228,246$ images, true captions and their corresponding negated captions. 
Using CC-Neg along with modifications to the 
contrastive loss of CLIP, our proposed CoN-CLIP framework, has an improved understanding of negations.
This training paradigm improves CoN-CLIP's ability to encode semantics reliably, resulting in $3.85\%$ average gain in top-1 accuracy for zero-shot image classification across 8 datasets.
Further, CoN-CLIP outperforms CLIP on challenging compositionality benchmarks such as SugarCREPE by $4.4\%$, showcasing emergent compositional understanding of objects, relations, and attributes in text.
Overall, our work addresses a crucial limitation of VLMs by introducing a dataset and framework that strengthens semantic associations between images and text, demonstrating improved large-scale foundation models with significantly reduced computational cost, promoting efficiency and accessibility. The code for our work can be found at: \textcolor{red}{\texttt{\href{https://github.com/jaisidhsingh/CoN-CLIP}{https://github.com/jaisidhsingh/CoN-CLIP}}}.

\end{abstract}
%
%
\section{Introduction}
\label{sec:introduction}
Generalised vision-language understanding is a crucial requirement for well-performing multimodal foundation models~\cite{lu2019vilbert, li2019visualbert, chen2020uniter, li2020oscar, jia2021scaling, clip, blip, flava}.
Contrastive learning has emerged as an effective way to learn a joint multimodal embedding space, by pulling representations of co-occurring images and texts close together and pushing those of unrelated pairs far away, based on their semantic and visual similarity~\cite{clip,blip}.
Vision-language models (VLMs)~\cite{clip, blip, flava, flamingo, omnivl, lit} use contrastive pretraining on web-scale image-annotation pairs~\cite{laion, cc12m} to achieve strong performance across a wide range of zero-shot baselines such as image-text matching, image retrieval, and object classification.
However, controlling the invariance in representations learned by these models is difficult as they depend upon the quality and diversity of training data, which makes generalising across unseen contexts a challenge~\cite{elephantintheroom}. Additionally, the contrastive learning objective optimises for the task of retrieval, which has been shown to encourage ``shortcut learning'', giving rise to models that behave like bag-of-words with an incomplete relational understanding of semantics~\cite{negclip}. 

\begin{figure*}[t]
    \centering
    \includegraphics[width=\linewidth]{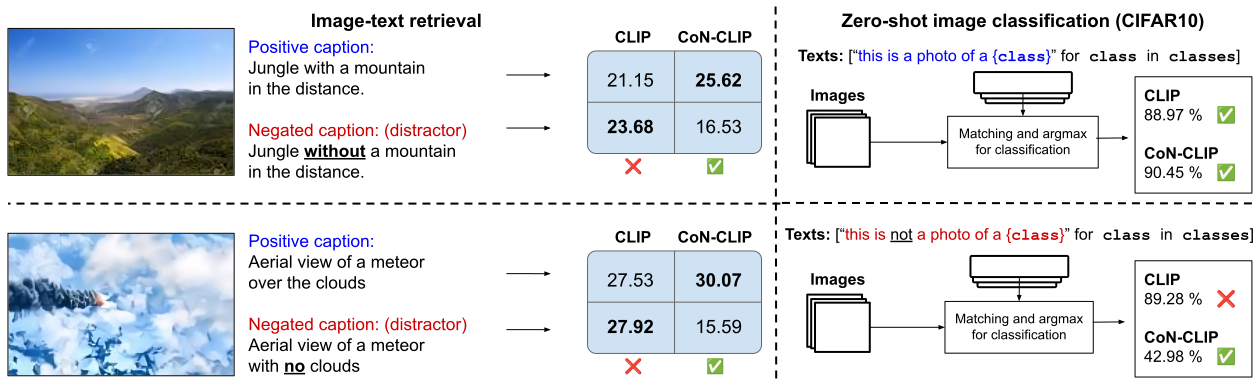}
    \caption{VLMs such as CLIP often match images to negation-based distractors with higher similarities than their true captions (left). Further, CLIP accurately retrieves images of a class even when prompted with ``this is \underline{not} a photo of a \{\texttt{class}\}'' (right).}
    \label{fig:intro_figure}
    \vspace{-10pt}
\end{figure*}
VLMs like CLIP~\cite{clip} often ignore negation words such as \textit{no, not,} and \textit{without}.
For example, an image of a dog matches with similar scores to both ``\textit{this is a photo of a dog}'' and ``\textit{this is \textbf{not} a photo of a dog}'' (Fig.~\ref{fig:intro_figure}). Also, the caption ``\textit{a car \textbf{without} tires}'' is frequently interpreted as an image of a car with tires (Fig.~\ref{fig:teaser}) by generative models using CLIP. 
Fig.~\ref{fig:intro_figure} illustrates further that VLMs inadequately capture negation words, indicating under-representation in the training data and a misalignment of negations with their correct implications in the image space.

Negations allow us to specify the absence of concepts~\cite{khemlani2014negations} and hence form an important part of logic and natural language.
However, negative sentences are harder to process than affirmative sentences~\cite{greco2020syntax, orenes2021looking}.
This is also highlighted through the under-representation of negatives in existing natural language inference benchmarks~\cite{hossain2020analysis, safavi2021negater} and that pretrained language models have difficulty performing well during neural translation tasks~\cite{hossain2020s} and fill-in-the-blank tests~\cite{kassner2020negated}. 
Understanding negations, though harder for learning-based models~\cite{testoni2022artificial}, is crucial for commonsense reasoning tasks~\cite{safavi2021negater, schon2021negation}.
This ability is highly desirable in image-text retrieval and text-to-image generation systems~\cite{stablediff} which employ VLMs.
To investigate this issue, we develop a comprehensive benchmark to evaluate VLMs' ability to understand explicit negations. 
We contribute the CC-Neg dataset, containing $228, 246$ image-caption pairs accompanied by grammatically correct and fluent \textit{negated captions}. The negated caption is a distractor text where a concept present in the image is negated explicitly using words such as \textit{no, not,} and \textit{without}.
We use the CC-3M dataset to generate (\textit{image, caption, negated caption}) triplets to test the negational understanding capabilities of VLMs. 
Through experiments on CC-Neg, we establish that VLMs generally do not understand prompts with negations and often match negated captions to the image over their true captions.

To mitigate this problem, we propose to augment the InfoNCE contrastive loss~\cite{infonce} with a fine-grained contrastive objective, by leveraging fluent and high-quality negated captions in CC-Neg and distractor images. 
CLIP's text encoder~\cite{clip} is fine-tuned using the proposed objective, and the resulting CoN-CLIP 
model shows drastic improvements on the negation-understanding task across varyingly complex negated captions.
Additionally, we find that our approach improves overall compositional understanding and outperforms CLIP by $4.4\%$ average R@1 on SugarCREPE, an unbiased benchmark for tasks such as replacing, adding, and swapping objects, attributes and relations in prompts.
This emphasises CoN-CLIP's ability to understand the semantic decomposition of scenes into objects and their association with various attributes and relations, without explicitly being trained with compositional prompts beyond negations.
Further, CoN-CLIP achieves notable improvements in top-1 zero-shot image-classification accuracy across $8$ datasets, namely ImageNet-1k~\cite{deng2009imagenet}, CIFAR-10~\cite{cifar10}, CIFAR-100~\cite{cifar10}, Caltech-101~\cite{caltech101}, Food-101~\cite{food101}, Flowers-102~\cite{flowers102}, Oxford Pets~\cite{oxfordpets}, and Stanford Cars~\cite{stanfordcars}, with the highest improvement being $10.95\%$ on CIFAR-100.
We summarise the contributions of this paper as follows:\vspace{-5pt}
\begin{enumerate}
\itemsep0em
    \item{We show that VLMs suffer in understanding negations in text and incorrectly associate them to images. For robust investigation of this phenomenon, we present CC-Neg, a large-scale dataset of $228, 246$ image-caption pairs along with high quality negated captions as distractors.}
    \item{Utilizing images and captions from the CC-Neg dataset, along with additional distractor images, we introduce CoN-CLIP, a fine-tuning framework that improves upon the original InfoNCE contrastive loss~\cite{infonce} and enhances the pretrained model's understanding of negations.}
    \item{Further, CoN-CLIP demonstrates enhanced performance on zero-shot image classification task and general purpose compositionality benchmarks, indicating a deeper understanding of visual concepts and improved compositional reasoning capabilities.}
\end{enumerate}
%
%
\section{Related Work}
\label{sec:related_work}
\subsubsection{Contrastive Image-Text Pretraining:}  
CLIP, one of the most popular VLMs, is contrastively pretrained on approximately 400M image-text pairs, and has emerged to be applicable for several tasks such as open-set attribute recognition~\cite{ovarnet} and object detection~\cite{owlvit}. New additions to CLIP's recipe such as image captioning with contrastive pretraining and self-supervision have produced models like BLIP~\cite{blip}, BLIP2~\cite{blip2}, SLIP~\cite{slip}. As a foundation model, CLIP has been used in image synthesis~\cite{stablediff, dalle2}, video-summarization~\cite{clipit}, and has led to the development of contrastive methods for modalities such as video~\cite{sdvideo} and audio~\cite{audioclip}. Our work evaluates how well such contrastively trained models capture compositionality by distinguishing the absence of concepts through negations.
\subsubsection{Compositional Understanding:} 
Towards compositional image-text matching,~\cite{comclip} presents a model to decompose images and texts into respective sub-images and words denoting subjects, objects, and predicates. Along similar lines,~\cite{negclip} presents ARO, a benchmark to study the sensitivity of VLMs to object order, relations, and attributes. The study shows that VLMs struggle with compositionality, and presents NegCLIP to improve on the investigated shortcomings.
Next, CREPE~\cite{crepe} presents a benchmark to evaluate compositionality in VLMs through systematicity and productivity. The systematicity component evaluates a VLM on seen and unseen contexts while productivity entails image-text matching with various types of hard negative captions which act as distractors while matching the image to its true caption. SugarCREPE~\cite{sugarcrepe} refines biases in CREPE and ARO to present a high-quality unbaised dataset. VLMs such as Neg-CLIP show significantly reduced improvements on SugarCREPE as compared to biased compositionality benchmarks like ARO and CREPE, implying an overfit on negative artifacts seen during training.
\subsubsection{Using Hard Negatives and Negations:} Hard negatives, or distractors which often lead to incorrect matching, are prominently used to evaluate image-text matching. CREPE and NegCLIP utilize such hard negatives to test sensitivity towards object order, swapping, relations, etc. Hard negatives are different from negations, which represent the absence of a concept. For instance, a simple negation is given by ``this is \textit{not} a cat'', which implies an object which does belong to the cat class. CLIPN~\cite{clipn} devises a method to learn prompts for CLIP which correspond to ``this is \textit{not} \texttt{X}''. These are used to perform out-of-distribution detection for CLIP. While our goal to teach CLIP to say no matches with CLIPN, we tackle the issue towards generalisable compositional understanding, beyond learning yes/no for simple prompt templates such as those employed in CLIPN.
\begin{table}[t]
    \caption{Comparison of CC-Neg with the Negate fold of CREPE-Productivity across true (\textit{P}) and negated (N) caption pairs. CC-Neg contributes a larger scale and greater diversity in the type of negation words used. Further, it exhibits greater fluency and plausibility in its text data as indicated by higher mean Vera scores~\cite{liu2023vera, sugarcrepe} for the negated captions ($0.347$ for CC-Neg versus $0.232$ for CREPE-Negate).}
    \label{tab:fluency_tab}
    \centering
    \begin{tabular*}{\linewidth}{@{\extracolsep{\fill}}ll}
    \toprule
        \textbf{\footnotesize{Dataset}} & \textbf{\footnotesize{Captions \textit{P} v/s N}} \\
        \midrule
        \multirow{4}{*}{\begin{tabular}{l}\footnotesize{CREPE}\\[\ExtraSep]\footnotesize{Negate}\end{tabular}} & {\footnotesize{\textit{P: Tree on a side of a street. street has on side a tree.}}}\\
         & {\footnotesize{N: Tree on a side of a street. Street not has on side a tree.}}\\
         \cmidrule{2-2}
         & {\footnotesize{\textit{P: Car has tires. There is a windows.}}}\\
         & {\footnotesize{N: There is no car has tire. There is a windows.}}\\
         \midrule
        \multirow{6}{*}{\footnotesize{CC-Neg}} & {\footnotesize{\textit{P: Festive banner with flags and an inscription.}}}\\
         & {\footnotesize{N: Festive banner with an inscription, but not with flags.}}\\
         \cmidrule{2-2}
         & {\footnotesize{\textit{P: A woman walks her dog on the beach.}}}\\
         & {\footnotesize{N: A woman walks on the beach without her dog.}}\\
         \cmidrule{2-2}
         & {\footnotesize{\textit{P: Dining table with kitchen in the background.}}}\\
         & {\footnotesize{N: Dining table with no kitchen.}}\\
        \bottomrule
        \\
    \end{tabular*}
    \vspace{-10pt}
\end{table}
\section{CC-Neg: Benchmark for Understanding of Negations}
\label{sec:ccneg}
Current datasets for image-text matching~\cite{cc12m, yfcc100m, thrush2022winoground, coco, desai2021redcaps} largely focus on matching images to their true captions in the presence of distractors (either distractor images or distractor texts). However, negations are rare in such datasets. 
The Negate fold of CREPE-Productivity~\cite{crepe} is an example dataset, with $17$K true image-caption pairs with $183$K distractor texts. The distractors include negation words, but suffer in terms of linguistic fluency (Table.~\ref{tab:fluency_tab}). 
This prevents the evaluation of negation understanding in VLMs in realistic settings. Hence, we introduce CC-Neg, a dataset aimed at comprehensively evaluating how well VLMs understand negations in realistic compositional prompts.
\subsection{CC-Neg: Benchmark for Evaluating VLMs on Negations}
\begin{figure*}[t]
    \centering
    \includegraphics[width=0.9\linewidth]{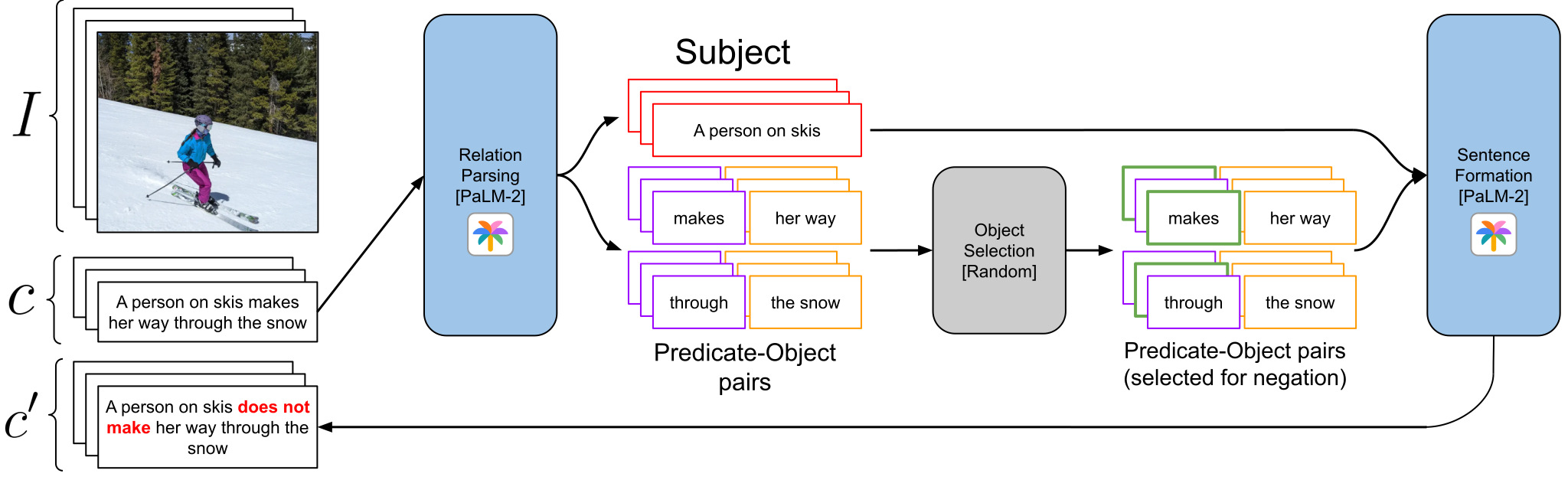}
    \caption{Overview of the generation of negated captions. Given the true caption of an image, an LLM (i) decomposes it into a \emph{subject} and \emph{predicate-object pairs}, and then (ii) selects a random predicate-object pair to negate to finally write the negated prompt.}
    \label{fig:ccneg_gen}
\end{figure*}

CC-Neg utilizes the Image-Labels subset, $300,000$ image-caption pairs, from the Conceptual Captions (CC-3M) dataset and a large language model (LLM) to obtain corresponding negated captions (overview in Fig.~\ref{fig:ccneg_gen}).
Given an image-caption pair $(I, c)$, we use PaLM-2~\cite{anil2023palm2} to generate a negated caption $c'$. For example, a true caption such as ``A city street with colorful billboards'' is used to write a negated caption ``A city street \textit{without} billboards''. 

More specifically, the negated caption $c'$ is obtained by prompting PaLM-2 to decompose $c$ into one \textit{subject}, and $\mathcal{K}$ \textit{predicate-object pairs} using in-context learning (ICL)~\cite{xie2022an, min2022rethinking}. Along with instructions to decompose the sentence into the above components, we add a handcrafted example within the prompt as a demonstration of the task. This allows PaLM-2 to effectively follow the schema required for this task. More details regarding the prompting method can be found in Sec.~\ref{sec:data_gen} of the supplementary material.
Next, for each caption, an object from the $\mathcal{K}$ pairs is randomly selected and its association to the subject as well as the scene is nullified using a negation word such as $\{not, no, without\}$ (Fig.~\ref{fig:ccneg_gen}). We use ICL in this step as well, where an example input and output is provided in the prompt as format. This results in the negated caption $c'$. In some cases, PaLM-2 does not faithfully decompose all captions. Additionally, PaLM-2 can negate objects by omission for certain samples. Such responses are considered erroneous and are excluded. Finally, CC-Neg contains $228, 246$ $(I, c, c')$ triplets.
We use CC-Neg 
as the test data for $4$ VLMs and evaluate their performance in associating true images with true captions in the following subsection.
\subsection{Evaluating VLMs on CC-Neg}
\label{ccneg_exp}
We benchmark four state-of-the-art VLMs: CLIP~\cite{clip}, BLIP~\cite{blip}, FLAVA~\cite{flava}, and Neg-CLIP~\cite{negclip} on CC-Neg, to test how well VLMs identify true image-caption pairing in the presence of negated captions as distractors.
\subsubsection{Experimental setup:} 
For each triplet $(I, c, c')$ in CC-Neg, a VLM computes a similarity-based match score $\phi(\cdot, \cdot)$ between each image-text pair. 
If $\phi(I, c) > \phi(I, c')$, the VLM is deemed to match the image $I$ to its true caption $c$ over the distractor and indicates a correct prediction. Alternately, $\phi(I, c) \leq \phi(I, c')$ signifies an incorrect prediction. Using this rule, we compute the accuracy of identifying true pairings for each VLM. For fair comparison of CLIP with Neg-CLIP, we use the ViT-B/32 architecture for both models. 
\begin{table}[h!]
\vspace{-10pt}
    \caption{For each VLM, we report the model and pretraining configurations alongside its accuracy on the entire CC-Neg dataset. \vspace{-4pt}}
    \label{tab:ccneg_caps_macro}
    \centering
    \begin{tabular*}{0.9\linewidth}{@{\extracolsep{\fill}}llc}
    \toprule
        \textbf{Model} & \textbf{Architecture \& Pretraining} & \textbf{CC-Neg Accuracy (\%)}\\
        \midrule
        CLIP & ViT-B/32 (OpenAI) & 66.4\\
        Neg-CLIP & ViT-B/32 (OpenAI+ARO fine-tuned) & 62.7\\
        FLAVA & Full (Meta) & 60.8\\
        BLIP & Base (Salesforce+COCO finetuned) & 63.5\\
        \bottomrule
        \\
    \end{tabular*}
    \vspace{-40pt}
\end{table}
\begin{figure}[t]
    \centering
    \includegraphics[width=\linewidth]{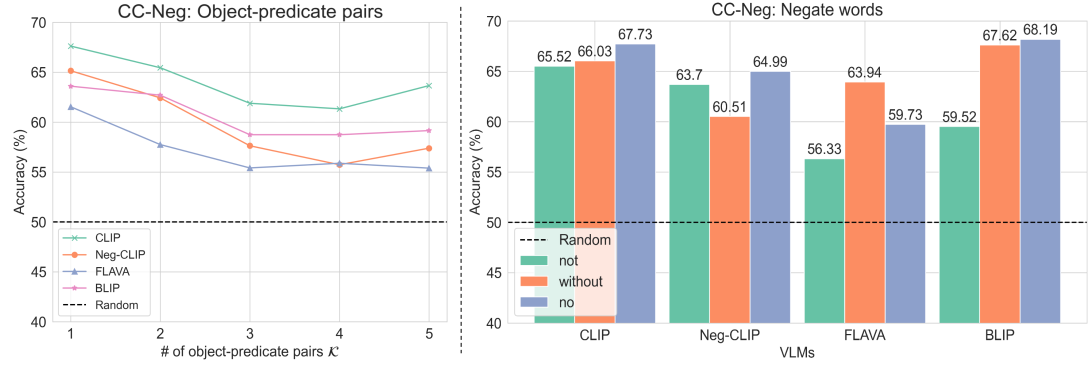}
    \caption{We report the accuracy of matching the image to its true caption for all VLMs, varying the number of predicate-objects, $\mathcal{K}$ from 1 to 5 (left). Additionally, we show the performance of all VLMs on each type of negation word used in CC-Neg (right).\vspace{-17pt}}
    \label{fig:ccneg_caps_num_ops_vlms}
\end{figure}
\subsubsection{Results:} The performance evaluation of state-of-the-art VLMs on CC-Neg results in the following observations.
\begin{enumerate}
\itemsep0em
    \item \textbf{VLMs fail to recognize negations:} We find that all VLMs exhibit poor understanding of negations in text.
The accuracy values in Table.~\ref{tab:ccneg_caps_macro} signify that VLMs often confuse negated captions as true ones. Specifically, the presence of the negated concept within $I$ is erroneously associated with $c'$, showing that VLMs largely ignore the effect of negation words like ``not'',``without'', etc.
Notably, Neg-CLIP, which otherwise outperforms CLIP on the Negate fold of CREPE~\cite{negclip}, does not show similar trends on CC-Neg. This can be attributed to our data generation procedure, where leveraging an LLM results in greater linguistic fluency in the negated captions. Consequently, our data domain differs from CREPE-Negate in the quality of distractor texts, which has more crude and non-fluent samples (shown in Table.~\ref{tab:fluency_tab}). This supports~\cite{sugarcrepe} in that Neg-CLIP exhibits biases towards non-fluent data. Overall, CLIP has the highest accuracy on CC-Neg, with Neg-CLIP, BLIP and FLAVA close but only slightly above random chance (50 \%).

\item \textbf{Performance degradation at higher complexities:} Next, we study the responses of the VLMs across all caption complexities (number of predicate-object pairs $\mathcal{K}$) in CC-Neg. Fig.~\ref{fig:ccneg_caps_num_ops_vlms} (left) depicts the accuracy of identifying true pairings for each value of $\mathcal{K}$. We find that models perform worse as the captions become more complex, arriving near random chance for all models except CLIP, supporting the claim that VLMs cannot compositionally understand negations. The presence of more objects and predicates likely obscures the effects of negation words and results in reduced performance. 

\item \textbf{VLMs favor certain negation words:} Lastly, we evaluate the effect of each negation word on the accuracy of a VLM. Fig.~\ref{fig:ccneg_caps_num_ops_vlms} (right) reports the accuracy of matching true pairs when the negation word in $c'$ is \textit{not}, \textit{without}, and \textit{no}. CLIP, Neg-CLIP, and BLIP are most accurate on \textit{no}, while FLAVA favors \textit{without}, reflected in its lower \textit{no} and \textit{not} accuracies.
\end{enumerate}

%
\section{Learning a Compositional Understanding of Negations}
\label{sec:conclip}
To address the limitations of VLMs in understanding negations and related concepts, we employ the proposed dataset CC-Neg to develop an improved contrastive framework, CoN-CLIP (Contrastive with Negations-CLIP).
Our proposed framework incorporates negated captions and relevant distractor images for fine-tuning, in addition to image and true caption pairs used in CLIP~\cite{clip}.
\subsection{Preliminaries}
\begin{figure}[t]
    \centering
    \includegraphics[width=0.8\linewidth]{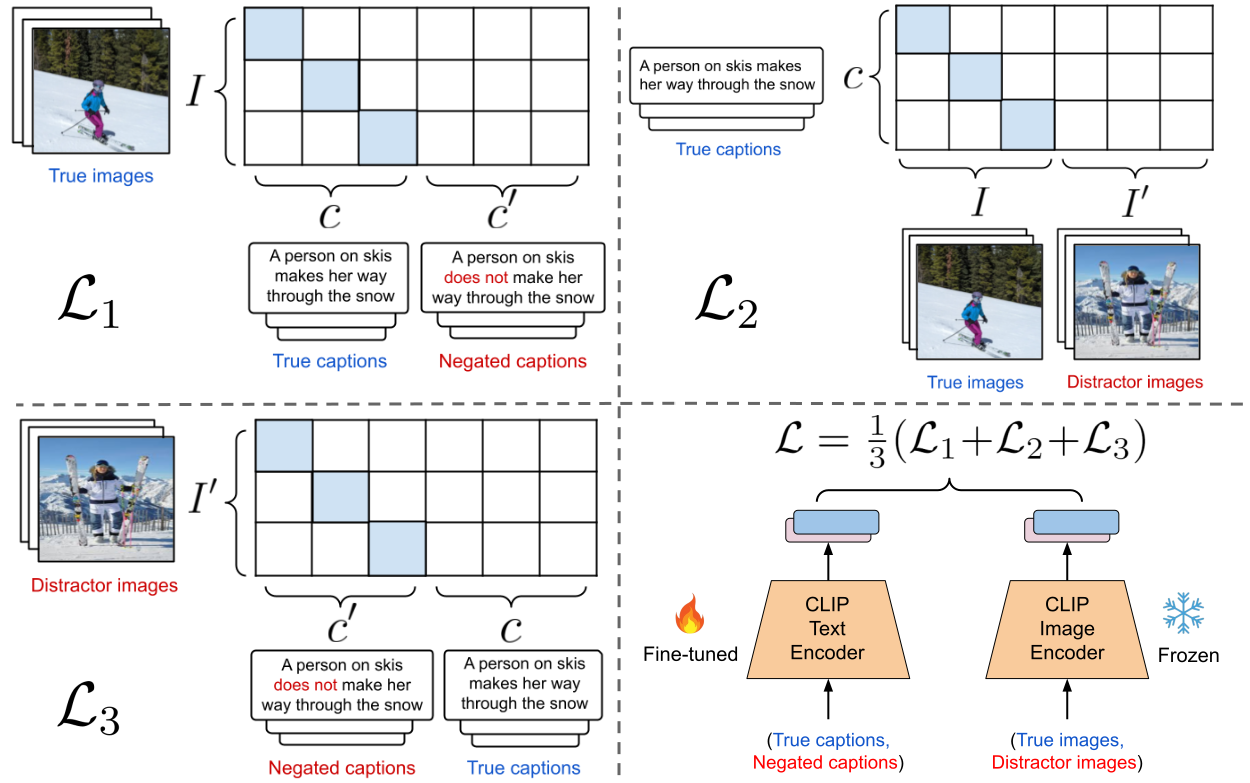}    
    \caption{We incorporate negations and distractor images in a contrastive objective for fine-tuning the CLIP text encoder towards improved negation understanding. The proposed loss functions are depicted above using the example of a training instance.}
    \label{fig:conclip_overview}
\end{figure}
The objective of CoN-CLIP is to allow CLIP or a similar VLM to observe negated captions and learn their implications on the scene's composition. This motivates the following design choices:
\begin{enumerate}
    \item \textbf{Negated captions per sample:} \hspace{2pt} We utilize a subset of CC-Neg, our large-scale dataset containing negation-based distractor texts. Specifically, the negated caption $c'_i$ is used alongside the true image-caption pair $(I_i, c_i)$.

    \item \textbf{Distractor images as reflections of negated captions:}\hspace{2pt} To anchor the effect of negations to visual concepts, we add distractor images which serve as \emph{crude} reflections of the negated caption $c'$. Providing visual context has shown to help model negation and its implications~\cite{testoni2022artificial}. Learning to repel such a distractor image $I'$ from the true caption $c$ shall lead to improved compositional awareness. 
    As an example, the distractor image corresponding to the caption ``A building in the sunset'' shall contain a building but not the sunset environment, in alignment with the negated caption. More details on obtaining distractor images is provided in Sec.~\ref{sec:distractor_images} of the supp. mat.
\end{enumerate}
Using negated captions and distractor images alongside the existing image-caption pairs, we compile a dataset $\mathcal{D} = \{I_i, c_i, c'_i, I'_i\}_{i=1}^N$. Here, $N$ is set to $188,246$ to hold out the remaining $40,000$ samples in CC-Neg for later evaluation. The next section describes contrastive learning used in CoN-CLIP.
\subsection{Fine-tuning CLIP with New Objectives}
Our modification of the contrastive objective of CLIP is given as follows. Let $f_{img}(\cdot)$ denote the image encoder and $f_{txt}(\cdot)$ the text encoder of CLIP. Using these encoders, we embed a set of $M$ images $\mathcal{I} = \{I_1, ... I_M \}$ and a set of $M$ captions $\mathcal{C} = \{c_1, ..., c_M \}$ to $E_{\mathcal{I}}$ and $E_{\mathcal{C}}$ respectively. Similarly, a set of negated captions $\mathcal{C}' = \{c'_1, ..., c'_M \}$ and a set of distractor images $\mathcal{I}' = \{I'_1, ... I'_M \}$ are embedded with their respective encoders to obtain 
$E_{\mathcal{C}'}$ and $E_{\mathcal{I}'}$. 
Here, each set of CLIP embedding belongs to $\mathbb{R}^{M \times d}$.
We then construct $3$ similarity matrices to be used in the final objective:
\begin{enumerate}
    \item \textbf{True images to all captions:}\hspace{1pt} $E_{\mathcal{C}}$ and $E_{\mathcal{C}'}$ are concatenated and the cosine-similarity of the concatenated caption embeddings with $E_{\mathcal{I}}$ are computed to obtain $P \in \mathbb{R}^{M \times 2M}$. The result of applying temperature $\tau$ and a column-wise softmax operation on $P$ is denoted by $\tilde{P}$. This computes loss $\mathcal{L}_1$ as
    \begin{equation}
        \mathcal{L}_1 = - \frac{1}{M} \sum_{i=1}^M \sum_{j=1}^{2M} \mathbbm{1}_{\{i = j\}} \hspace{1pt} \log(\tilde{P}_{ij})
    \end{equation}

    \item \textbf{True captions to all images:}\hspace{1pt} In this step, $E_{\mathcal{I}}$ and $E_\mathcal{I}'$ are concatenated to compute their cosine-similarity with $E_\mathcal{C}$. The resultant similarity matrix is denoted by $Q \in \mathbb{R}^{M \times 2M}$. Next, we obtain $\tilde{Q}$ by applying temperature $\tau$ and a column-wise softmax on $Q$ to compute loss $\mathcal{L}_2$ as
    \begin{equation}
        \mathcal{L}_2 = - \frac{1}{M} \sum_{i=1}^M \sum_{j=1}^{2M} \mathbbm{1}_{\{i = j\}} \hspace{1pt} \log(\tilde{Q}_{ij})
    \end{equation}
    \item \textbf{Distractor images to all captions:} Here, $E_{\mathcal{C}'}$ and $E_\mathcal{C}$ are concatenated after which the cosine-similarity matrix between $E_\mathcal{I'}$ and the concatenated image embeddings is computed as $R \in \mathbb{R}^{M \times 2M}$. The matrix $R$ is subsequently scaled by $\tau$ and column-wise softmaxed to give $\tilde{R}$, computing $\mathcal{L}_3$ as
    \begin{equation}
        \mathcal{L}_3 = - \frac{1}{M} \sum_{i=1}^M \sum_{j=1}^{2M} \mathbbm{1}_{\{i = j\}} \hspace{1pt} \log(\tilde{R}_{ij})
    \end{equation}
\end{enumerate}
Finally, we compute the total loss $\mathcal{L}_{conclip}$ as $\mathcal{L}_{conclip} = \frac{1}{3}(\mathcal{L}_1+\mathcal{L}_2+\mathcal{L}_3)$. Observing the lack of understanding of negations in text, it becomes necessary to train the embedding layer and the attention mechanisms of the text encoder. This is done to impart new knowledge of how negations affect the semantics of the given scene. Hence, we freeze the image encoder
and fine-tune CLIP's text encoder on our final loss function $\mathcal{L}_{conclip}$, similar to~\cite{lit}. 
The learning rate is initialized as $1e-6$ which follows a cosine schedule of $50$ warmup steps. The optimizer used is AdamW~\cite{adamw} with $0.2$ weight decay and a batch of size $256$. We use PyTorch~\cite{paszke2019pytorch} for all experiments which are run on one NVIDIA V100 GPU.
\section{Experiments}
\label{sec:experiments}
This section describes experiments evaluating the understanding of the proposed framework for various tasks and its comparison with existing techniques. Firstly, we evaluate CoN-CLIP's ability to understand negations in Sec.~\ref{sec:negations}. Secondly, Sec.~\ref{sec:zeroshot} investigates how the proposed modifications to the contrastive loss affects zero-shot image classification. Further, Sec.~\ref{sec:gencomp} tests the emerging decomposition of concepts learned by CoN-CLIP on tasks for general-purpose compositionality.
In the following, CoN-CLIP denotes CLIP fine-tuned on CC-Neg using all losses ($\mathcal{L}_{conclip}$) unless explicitly specified.
\subsection{Understanding of Negations}
\label{sec:negations}
\subsubsection{Experimental setup:} CoN-CLIP is compared with other VLMs mentioned in Sec.~\ref{ccneg_exp} using the ViT-B/32 backbone across all CLIP-based models for fair comparison.
To test our framework's understanding of negations, we use a held-out evaluation set from CC-Neg containing $40,000$ $(I, c, c')$ triplets to measure the accuracy of matching image $I$ to the true caption $c$ in the presence of the negated caption $c'$ as explained in Sec.~\ref{ccneg_exp}.
As mentioned in Sec.~\ref{sec:ccneg}, CREPE is another dataset containing negations. However, it is unsuitable for evaluating CoN-CLIP due to its biased nature as well as non-fluent distractor captions~\cite{sugarcrepe}.\vspace{-8pt}
\begin{table}[h]
    \caption{Evaluating CoN-CLIP alongside all chosen VLMs on CC-Neg. \underline{Underlined} values denote highest performance across all models.}
    \label{tab:conclip_ccneg_macro}
    \vspace{-10pt}
    \centering
    \begin{tabular*}{\linewidth}{@{\extracolsep{\fill}}llc}
    \toprule
        \multirow{2}{*}{\textbf{Model}} & \textbf{Architecture \&} & \textbf{CC-Neg}\\
         & \textbf{Pretraining} & \textbf{Accuracy (\%) $\uparrow$}\\
        \midrule
        CLIP & ViT-B/32 (OpenAI) & 65.70\\
        Neg-CLIP & ViT-B/32 (OpenAI+ARO fine-tuned) & 62.63\\
        FLAVA & Full (Meta) & 58.93\\
        BLIP & Base (Salesforce+COCO fine-tuned) & 62.31\\
        CoN-CLIP & ViT-B/32 (OpenAI+CC-Neg fine-tuned) & \underline{99.70}\\
        \bottomrule
        \\
    \end{tabular*}
    \vspace{-30pt}
\end{table}
\subsubsection{Results:}
Matching accuracy of CoN-CLIP on CC-Neg evaluation set is reported in Table.~\ref{tab:conclip_ccneg_macro} alongside CLIP, Neg-CLIP, FLAVA, and BLIP. CoN-CLIP outperforms other VLMs by a large margin ($>30 \%$) on the held-out samples for all caption complexities (number of predicate-object pairs $\mathcal{K}$). While CoN-CLIP's performance decreases as the value of $\mathcal{K}$ increases, the drop in performance is significantly less and does not fall below $99\%$ even for $\mathcal{K}=5$.
Similarly, CoN-CLIP improves in performance across each type of negation word. Here, CoN-CLIP performs the worst for negated captions containing \textit{no} as the negation word ($96.5\%$), while still outperforming other VLMs (best being BLIP at $68.19\%$) on such samples. These results are provided in further detail in Sec.~\ref{sec:more_results} of the supp. mat.
These results show that CoN-CLIP exhibits a greater understanding of negations in text as compared to other VLMs. Further, it learns to reliably reject captions which negate visually-present concepts.

Additionally, we evaluate if CoN-CLIP can transfer its understanding to prompts which directly negate the subject of the text. For this, we use $8$ popular benchmarks for image classification. Following the example in Fig.~\ref{fig:intro_figure}, image classification accuracy is computed using two types of class prompts: ``this is a photo of a \{\texttt{class}\}'' (standard), and ``this is \underline{not} a photo of a \{\texttt{class}\}'' (negated). The latter must be matched to images which do not belong to the ``\texttt{class}'' category, indicated in low top-1 accuracy. To benchmark this behavior, we compute $\Delta$, the difference between top-1 accuracies obtained by using standard class prompts and those obtained by using negated class prompts. This $\Delta$ value is computed for $8$ image classification datasets, namely ImageNet-1k~\cite{deng2009imagenet}, CIFAR-10~\cite{cifar10}, CIFAR-100~\cite{cifar10}, Caltech-101~\cite{caltech101}, Food-101~\cite{food101}, Flowers-102~\cite{flowers102}, Oxford Pets~\cite{oxfordpets}, and Stanford Cars~\cite{stanfordcars}, and averaged to obtain a single measure.

It is desirable to show high accuracy while using standard class prompts, however, negated prompts for a given class must show low accuracy.
CoN-CLIP is able to correctly reject images when observing
negated class prompts indicated in the significantly higher mean $\Delta$ for CoN-CLIP (62.03\%) versus that of CLIP (0.98\%).
This shows the ability of CoN-CLIP to generalise to subject negations and correctly identify concepts to reject beyond its training data.
\subsection{Zero-shot Image Classification}
\label{sec:zeroshot}
The proposed framework addresses the limitations in understanding negations and their visual associations. However, it is also necessary to investigate how CoN-CLIP fine-tuning impacts the capabilities of CLIP across various tasks and domains. Here, we evaluate CoN-CLIP on zero-shot image classification, a task where CLIP shows strong performance across several datasets. \vspace{-10pt}
\subsubsection{Experimental setup:}  We evaluate the effect of our fine-tuning process across all CLIP architectures ViT-B/16, ViT-B/32, and ViT-L/14 which are also used as baselines for comparison. We use the same 8 image classification benchmarks mentioned in Sec.~\ref{sec:negations} to measure top-$1$ zero-shot classification accuracy.\vspace{-10pt}
\begin{table*}[h!]
    \caption{Evaluation of CoN-CLIP on zero-shot image classification shows improvements across all datasets. Here, highest accuracy values for a dataset are \underline{underlined}, while highest accuracy values for a CLIP backbone are given in \textit{italics}.}
    \label{tab:zeroshot_table}
    \centering
    \begin{tabular*}{\linewidth}{@{\extracolsep{\fill}}lcccccccc}
        \toprule
        \cmidrule{2-9}
        \multirow{2}{*}{\textbf{Model}} & \textbf{ImageNet} & \textbf{Caltech} & \textbf{Flowers} & \textbf{CIFAR} & \textbf{Food} & \textbf{Stanford} & \textbf{Oxford} & \textbf{CIFAR} \\
         & \textbf{1k} & \textbf{101} & \textbf{102} & \textbf{100} & \textbf{101} & \textbf{Cars} & \textbf{Pets} & \textbf{10} \\
        \midrule 
        \small{\textbf{CLIP}} & & & & & & & & \\
        \small{ViT-B/16} & \small{68.35} & \small{82.56} & \small{64.14} & \small{53.54} & \small{86.89} & \small{61.68} & \small{81.82} & \small{88.23} \\
        \small{ViT-B/32} & \small{63.36} & \small{81.50} & \small{60.50} & \small{55.18} & \small{81.15} & \small{58.33} & \small{80.08} & \small{88.97}\\
        \small{ViT-L/14} & \small{75.51} & \small{81.80} & \small{72.42} & \small{65.95} & \small{92.10} & \small{74.64} & \small{88.06} & \small{91.40}\\
        \midrule  
        \small{\textbf{CoN-CLIP}} & & & & & & & & \\
        \small{ViT-B/16} & \textit{68.95} & \textit{87.62} & \textit{66.69} & \textit{64.49} & \textit{88.13} & \textit{62.08} & \textit{85.45} & \textit{90.88} \\ 
        \small{ViT-B/32} & \textit{63.36} & \textit{86.91} & \textit{64.74} & \textit{62.31} & \textit{83.39} & \textit{58.84} & \textit{81.66} & \textit{90.45}\\
        \small{ViT-L/14} & \small{\underline{\textit{75.93}}} & \small{\underline{\textit{87.90}}} & \small{\underline{\textit{75.12}}} & \small{\underline{\textit{75.39}}} & \small{\underline{\textit{93.01}}} & \small{\underline{\textit{76.17}}} & \small{\underline{\textit{89.32}}} & \small{\underline{\textit{95.05}}}\\
        \bottomrule
        \\
    \end{tabular*}
    \vspace{-15pt}
\end{table*}
\begin{figure}[h!]
    \centering
    \includegraphics[width=\linewidth]{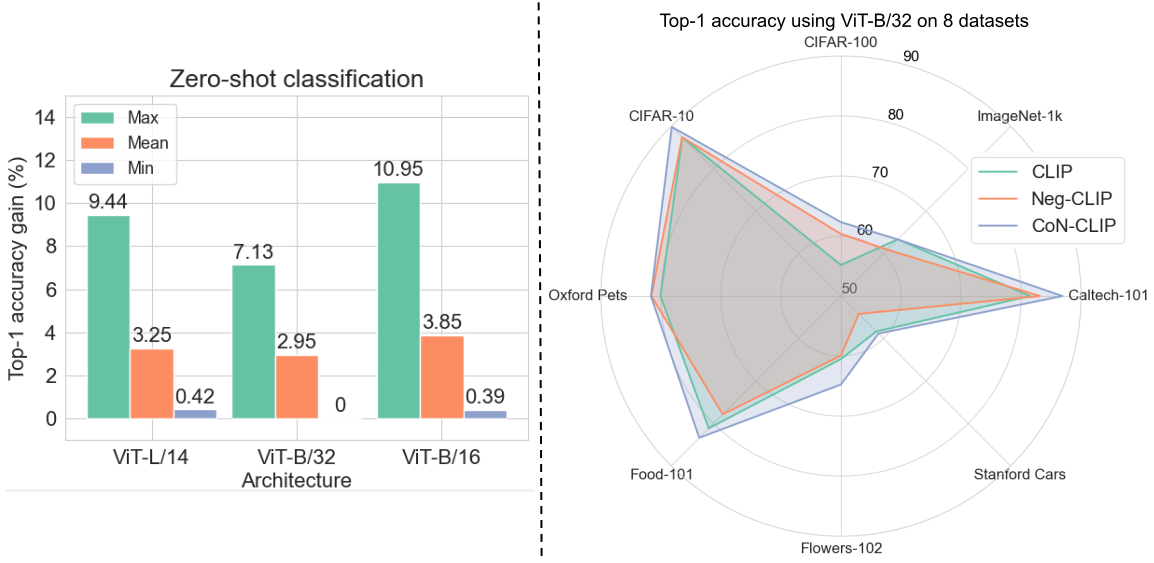}
    \caption{We present performance gains of CoN-CLIP over CLIP across all datasets per architecture (left). Additionally, we compare CoN-CLIP on image classification with Neg-CLIP using the ViT-B/32 backbone (right).}
    \label{fig:zeroshot_radar}
    \vspace{-17pt}
\end{figure}
\subsubsection{Results:} Table.~\ref{tab:zeroshot_table} presents a comprehensive evaluation of all model architectures on all datasets mentioned above. Considering CLIP as baseline, we find that CoN-CLIP shows greater or equal top-1 accuracy for all datasets and architecture. As shown in Fig.~\ref{fig:zeroshot_radar}, CoN-CLIP ViT-B/16 exhibits an average improvement of $3.85\%$ across all datasets, with the highest improvement of $10.95\%$ on the CIFAR-100 dataset. Overall, CoN-CLIP presents an average gain of $3.19\%$ in top-1 accuracy across all datasets and architectures. Additionally, we also use Neg-CLIP as a baseline for the ViT-B/32 architecture and present its performance alongside CoN-CLIP in Fig.~\ref{fig:zeroshot_radar}. We find that Neg-CLIP falls below CLIP ViT-B/32 on top-1 accuracy when evaluated on ImageNet-1k, Stanford Cars, Flowers-102, Food-101. This validates that our fine-tuning process positively impacts CLIP's understanding of negations as well as zero-shot classification. By learning to disentangle the effect of negation words, CoN-CLIP reduces bag of words-like behavior and allows semantics to align better.
\subsection{General Purpose Compositional Understanding}
\label{sec:gencomp}
To understand a scene as a function of its individual components, a model must learn to parse object relations and attributes. Thus, it is necessary to evaluate CoN-CLIP on data domains specifically designed to benchmark fine-grained compositional understanding. This experiment evaluates the performance of CoN-CLIP on tasks pertaining to attributes and relations in natural scenes. Such an evaluation of generalisability in a different data domain aims to show that learning negations can strengthen overall compositional understanding.\vspace{-10pt}
\subsubsection{Experimental setup:} We evaluate CoN-CLIP on SugarCREPE and use CLIP as baseline for zero-shot image-text matching. Specifically, a VLM must match a given image to its true caption by correctly rejecting the provided false caption which may contain replaced/added/swapped objects, attributes and relations. Notably, we test CoN-CLIP on SugarCREPE without fine-tuning it on any additional data tailored towards compositionality.\vspace{-10pt}
\subsubsection{Results:} We present the results of this experiment in Table.~\ref{tab:sugarcrepe}.
Out of $21$ total settings, CoN-CLIP outperforms CLIP in $18$ settings on SugarCREPE, showing an average improvement of $4.4\%$ in retrieval performance (R@1). 
In particular, the largest improvements are for the Add fold of SugarCREPE where the average gain in retrieval accuracy is $10.65\%$ for Add-Object, and $9.64\%$ for Add-Attribute.
We infer that this occurs due to the implicit effects of the proposed fine-tuning process. Considering that negated captions are essentially fine-grained variations of the true captions, learning to repel negated captions in the proposed objective increases the sensitivity of the model to changes in the atoms of input texts.
Moreover, it allows CoN-CLIP to pay greater attention to how concepts are composed in text by forcing the model to prioritise associations with correctly composed semantics. This resolves CLIP's initial tendency to naively match images to the mere presence of semantics in text.
\vspace{-20pt}
\begin{table}[h!]
    \caption{Evaluating CoN-CLIP on SugarCREPE alongside CLIP on R@1. Highest performance for a fold and CLIP backbone are \underline{underlined} and \textit{italicised} respectively.}
    \label{tab:sugarcrepe}
    \centering
    \begin{tabular*}{0.9\linewidth}{@{\extracolsep{\fill}}lccccccc}
       \toprule
       \multirow{2}{*}{\textbf{Model}}  &  \multicolumn{3}{c}{\textbf{Replace}} & \multicolumn{2}{c}{\textbf{Add}} & \multicolumn{2}{c}{\textbf{Swap}}\\
        \cmidrule{2-4}\cmidrule{5-6}\cmidrule{7-8}
        & {Object} & {Attribute} & {Relation} & {Object} & {Attribute} & {Object} & {Attribute}\\
        \midrule
        \textbf{CLIP} & & & & & & & \\
        ViT-B/16 & 93.28 & 80.83 & \textit{66.00} & 78.32 & 66.61 & \textit{59.59} & 64.41\\
        ViT-B/32 & 90.79 & 80.07 & \textit{68.99} & 76.91 & 68.35 & 60.81 & 63.06\\
        ViT-L/14 & 94.06 & 79.18 & 65.07 & 78.17 & 71.38 & 60.00 & 62.16\\
        \midrule
        \textbf{CoN-CLIP} & & & & & & & \\
        ViT-B/16 & \textit{93.58} & \textit{80.96} & 63.30 & \textit{87.29} & \textit{79.62} & 59.18 & \textit{65.16} \\
        ViT-B/32 & \textit{91.76} & \textit{80.96} & 66.28 & \textit{87.92} & \textit{78.03} & \textit{63.67} & \textit{66.96}\\
        ViT-L/14 & \textit{\underline{95.31}} & \textit{\underline{81.72}} & \textit{\underline{66.99}} & \textit{\underline{90.15}} & \textit{\underline{77.60}} & \textit{\underline{65.36}} & \textit{\underline{63.06}} \\
    \bottomrule
    \\
    \end{tabular*}
    \vspace{-15pt}
\end{table}
\subsection{Ablation Study}
\label{sec:ablation}
We conduct an ablation study to understand the effect of each loss function and the results are shown in Table~\ref{tab:ablation}.
Specifically, we report average R@1 for each fold of SugarCREPE. Next, the accuracy of matching an image to its true caption is given for the CC-Neg dataset, and the average top-1 image classification accuracy across all datasets used in Sec.~\ref{sec:zeroshot}.
Additionally, we study the effect of various contrastive loss design choices, and the effect of choosing between CC-Neg and CREPE-Negate data domains.
Fine-tuning CLIP on CREPE-Negate with $\mathcal{L}_1$ (using CREPE hard negatives as $c'$), results in significantly lower performance (refer Table~\ref{tab:ablation}) across all tasks and data domains. This validates our design choices for fine-tuning and the proposed dataset CC-Neg. \vspace{-10pt}
\begin{table}[h!]
    \vspace{-5pt}
    \caption{Our ablation study with the CLIP ViT-B/32 backbone and different combinations of loss terms across all experiments (negation understanding, image classification, general compositionality). CC-Neg - $\mathcal{L}_{conclip}$ yields the highest average performance (\underline{underlined}) across all settings, strongly outperforming CREPE-Negate - $\mathcal{L}_{1}$.}
    \label{tab:ablation}
    \centering
    \begin{tabular*}{\linewidth}{@{\extracolsep{\fill}}lccccc}
        \toprule
        \multirow{2}{*}{\textbf{Dataset - Loss}} & \multicolumn{3}{c}{\textbf{SugarCREPE R@1}} & \textbf{CC-Neg} & \textbf{Image classification}\\
         &  \textbf{Replace} & \textbf{Add} & \textbf{Swap} & \textbf{Accuracy} & \textbf{Top-1 accuracy} \\  
        \cmidrule{1-1}\cmidrule{2-4}\cmidrule{5-5}\cmidrule{6-6}
        CC-Neg - $\mathcal{L}_1$ & 79.36 & 82.22 & 61.64 & 99.76 & 73.32\\
        CC-Neg - $\mathcal{L}_2$ & 79.38 & 85.26 & 65.88 & 56.07 & 72.97\\
        CC-Neg - $\mathcal{L}_{1}+\mathcal{L}_2$ & 80.55 & 83.29 & 64.07 & 99.72 & 73.37\\
        CC-Neg - $\mathcal{L}_{conclip}$ & \underline{79.67} & \underline{82.97} & \underline{65.18} & \underline{99.70} & \underline{73.95}\\
        \midrule
        CREPE-Negate - $\mathcal{L}_1$ & 72.40 & 81.14 & 61.94 & 69.79 & 70.55 \\
        \bottomrule
        \\
    \end{tabular*}
    \vspace{-30pt}
\end{table}
\section{Discussion and Conclusion}
\label{sec:conclusion}
This paper investigates how well VLMs are able to understand negations in text. Particularly, we find that VLMs often ignore negations in text and incorrectly match them to images depicting the negated concept.
Addressing the need for fluent text data containing negations, we present CC-Neg, a new dataset for the evaluation of negation understanding, leveraging LLMs for mining hard negative texts. 
Further, CoN-CLIP, a framework aimed at mitigating this problem is proposed. This is achieved by supplementing the contrastive learning process with captions containing negations, as well as distractor images. Our results show that CoN-CLIP, with the help of examples from CC-Neg, is able to understand negations significantly better than CLIP and other similar VLMs such as Neg-CLIP, FLAVA, and BLIP.
The proposed objective enhances CoN-CLIP's sensitivity to the composition of atoms in text, fostering emergent and general compositional understanding, evidenced by its performance on non-negation tasks.

Our work uncovers an important approach to address compositionality by indicating how concepts are composed through negations. Notably, the proposed framework is aimed beyond benchmarking purposes, rather, we aim to inspire effective data-driven algorithmic improvements to foundation models. We constructively provide a foundation model with knowledge of concepts rarely seen during pretraining, without requiring the scale of the same. This work motivates flexibility and feasibility in foundation model research, which is often unapproachable to researchers with limited computational resources. 
\bibliographystyle{splncs04}
\newpage\bibliography{main}

\begin{thebibliography}{10}
\providecommand{\url}[1]{\texttt{#1}}
\providecommand{\urlprefix}{URL }
\providecommand{\doi}[1]{https://doi.org/#1}

\bibitem{midjourney}
\url{https://www.midjourney.com/}

\bibitem{flamingo}
Alayrac, J.B., Donahue, J., Luc, P., Miech, A., Barr, I., Hasson, Y., Lenc, K., Mensch, A., Millican, K., Reynolds, M., Ring, R., Rutherford, E., Cabi, S., Han, T., Gong, Z., Samangooei, S., Monteiro, M., Menick, J.L., Borgeaud, S., Brock, A., Nematzadeh, A., Sharifzadeh, S., Bi\'{n}kowski, M.a., Barreira, R., Vinyals, O., Zisserman, A., Simonyan, K.: Flamingo: a visual language model for few-shot learning. In: Koyejo, S., Mohamed, S., Agarwal, A., Belgrave, D., Cho, K., Oh, A. (eds.) Advances in Neural Information Processing Systems. vol.~35, pp. 23716--23736. Curran Associates, Inc. (2022)

\bibitem{anil2023palm2}
Anil, R., Dai, A.M., Firat, O., Johnson, M., Lepikhin, D., Passos, A., Shakeri, S., Taropa, E., Bailey, P., Chen, Z., et~al.: Palm 2 technical report. arXiv preprint arXiv:2305.10403  (2023)

\bibitem{dalle3}
Betker, J., Goh, G., Jing, L., Brooks, T., Wang, J., Li, L., Ouyang, L., Zhuang, J., Lee, J., Guo, Y., et~al.: Improving image generation with better captions. Computer Science. https://cdn. openai. com/papers/dall-e-3. pdf (3), ~8 (2023)

\bibitem{food101}
Bossard, L., Guillaumin, M., Van~Gool, L.: Food-101--mining discriminative components with random forests. In: European Conference on Computer Vision. pp. 446--461. Springer (2014)

\bibitem{sdvideo}
Chai, W., Guo, X., Wang, G., Lu, Y.: Stablevideo: Text-driven consistency-aware diffusion video editing. In: Proceedings of the IEEE/CVF International Conference on Computer Vision. pp. 23040--23050 (2023)

\bibitem{cc12m}
Changpinyo, S., Sharma, P., Ding, N., Soricut, R.: Conceptual 12m: Pushing web-scale image-text pre-training to recognize long-tail visual concepts. In: Proceedings of the IEEE/CVF Conference on Computer Vision and Pattern Recognition. pp. 3558--3568 (2021)

\bibitem{ovarnet}
Chen, K., Jiang, X., Hu, Y., Tang, X., Gao, Y., Chen, J., Xie, W.: Ovarnet: Towards open-vocabulary object attribute recognition. In: Proceedings of the IEEE/CVF Conference on Computer Vision and Pattern Recognition. pp. 23518--23527 (2023)

\bibitem{chen2020uniter}
Chen, Y.C., Li, L., Yu, L., El~Kholy, A., Ahmed, F., Gan, Z., Cheng, Y., Liu, J.: Uniter: Universal image-text representation learning. In: European Conference on Computer Vision. pp. 104--120. Springer (2020)

\bibitem{deng2009imagenet}
Deng, J., Dong, W., Socher, R., Li, L.J., Li, K., Fei-Fei, L.: Imagenet: A large-scale hierarchical image database. In: Proceedings of the IEEE/CVF Conference on Computer Vision and Pattern Recognition. pp. 248--255. IEEE (2009)

\bibitem{desai2021redcaps}
Desai, K., Kaul, G., Aysola, Z., Johnson, J.: Redcaps: Web-curated image-text data created by the people, for the people. arXiv preprint arXiv:2111.11431  (2021)

\bibitem{laclip}
Fan, L., Krishnan, D., Isola, P., Katabi, D., Tian, Y.: Improving clip training with language rewrites. Advances in Neural Information Processing Systems  \textbf{36} (2024)

\bibitem{caltech101}
Fei-Fei, L., Fergus, R., Perona, P.: Learning generative visual models from few training examples: An incremental bayesian approach tested on 101 object categories. In: Conference on Computer Vision and Pattern Recognition Workshop. pp. 178--178. IEEE (2004)

\bibitem{greco2020syntax}
Greco, M.: On the syntax of surprise negation sentences: A case study on expletive negation. Natural Language \& Linguistic Theory pp. 775--825 (2020)

\bibitem{audioclip}
Guzhov, A., Raue, F., Hees, J., Dengel, A.: Audioclip: Extending clip to image, text and audio. In: IEEE International Conference on Acoustics, Speech and Signal Processing (ICASSP). pp. 976--980. IEEE (2022)

\bibitem{hossain2020s}
Hossain, M.M., Anastasopoulos, A., Blanco, E., Palmer, A.: It’s not a non-issue: Negation as a source of error in machine translation. In: Proceedings of the 2020 Conference on Empirical Methods in Natural Language Processing. pp. 3869--3885 (2020)

\bibitem{hossain2020analysis}
Hossain, M.M., Kovatchev, V., Dutta, P., Kao, T., Wei, E., Blanco, E.: An analysis of natural language inference benchmarks through the lens of negation. In: Proceedings of the 2020 Conference on Empirical Methods in Natural Language Processing. pp. 9106--9118 (2020)

\bibitem{sugarcrepe}
Hsieh, C.Y., Zhang, J., Ma, Z., Kembhavi, A., Krishna, R.: Sugarcrepe: Fixing hackable benchmarks for vision-language compositionality. Advances in Neural Information Processing Systems  (2024)

\bibitem{jia2021scaling}
Jia, C., Yang, Y., Xia, Y., Chen, Y.T., Parekh, Z., Pham, H., Le, Q., Sung, Y.H., Li, Z., Duerig, T.: Scaling up visual and vision-language representation learning with noisy text supervision. In: International Conference on Machine Learning. pp. 4904--4916. PMLR (2021)

\bibitem{comclip}
Jiang, K., He, X., Xu, R., Wang, X.E.: Comclip: Training-free compositional image and text matching. arXiv preprint arXiv:2211.13854  (2022)

\bibitem{kassner2020negated}
Kassner, N., Sch{\"u}tze, H.: Negated and misprimed probes for pretrained language models: Birds can talk, but cannot fly. In: Proceedings of the 58th Annual Meeting of the Association for Computational Linguistics. pp. 7811--7818 (2020)

\bibitem{khemlani2014negations}
Khemlani, S., Orenes, I., Johnson-Laird, P.N.: The negations of conjunctions, conditionals, and disjunctions. Acta Psychologica pp.~1--7 (2014)

\bibitem{stanfordcars}
Krause, J., Stark, M., Deng, J., Fei-Fei, L.: 3d object representations for fine-grained categorization. In: Proceedings of the IEEE International Conference on Computer Vision Workshops. pp. 554--561 (2013)

\bibitem{cifar10}
Krizhevsky, A., Hinton, G., et~al.: Learning multiple layers of features from tiny images  (2009)

\bibitem{blip2}
Li, J., Li, D., Savarese, S., Hoi, S.: Blip-2: Bootstrapping language-image pre-training with frozen image encoders and large language models. arXiv preprint arXiv:2301.12597  (2023)

\bibitem{blip}
Li, J., Li, D., Xiong, C., Hoi, S.: Blip: Bootstrapping language-image pre-training for unified vision-language understanding and generation. In: International Conference on Machine Learning. pp. 12888--12900. PMLR (2022)

\bibitem{li2019visualbert}
Li, L.H., Yatskar, M., Yin, D., Hsieh, C.J., Chang, K.W.: Visualbert: A simple and performant baseline for vision and language. arXiv preprint arXiv:1908.03557  (2019)

\bibitem{li2020oscar}
Li, X., Yin, X., Li, C., Zhang, P., Hu, X., Zhang, L., Wang, L., Hu, H., Dong, L., Wei, F., et~al.: Oscar: Object-semantics aligned pre-training for vision-language tasks. In: European Conference on Computer Vision. pp. 121--137. Springer (2020)

\bibitem{coco}
Lin, T.Y., Maire, M., Belongie, S., Hays, J., Perona, P., Ramanan, D., Doll{\'a}r, P., Zitnick, C.L.: Microsoft coco: Common objects in context. In: European Conference on Computer Vision. pp. 740--755. Springer (2014)

\bibitem{liu2023vera}
Liu, J., Wang, W., Wang, D., Smith, N.A., Choi, Y., Hajishirzi, H.: Vera: A general-purpose plausibility estimation model for commonsense statements. arXiv preprint arXiv:2305.03695  (2023)

\bibitem{adamw}
Loshchilov, I., Hutter, F.: Decoupled weight decay regularization. arXiv preprint arXiv:1711.05101  (2017)

\bibitem{lu2019vilbert}
Lu, J., Batra, D., Parikh, D., Lee, S.: Vilbert: Pretraining task-agnostic visiolinguistic representations for vision-and-language tasks. Advances in Neural Information Processing Systems  (2019)

\bibitem{crepe}
Ma, Z., Hong, J., Gul, M.O., Gandhi, M., Gao, I., Krishna, R.: Crepe: Can vision-language foundation models reason compositionally? In: Proceedings of the IEEE/CVF Conference on Computer Vision and Pattern Recognition. pp. 10910--10921 (2023)

\bibitem{min2022rethinking}
Min, S., Lyu, X., Holtzman, A., Artetxe, M., Lewis, M., Hajishirzi, H., Zettlemoyer, L.: Rethinking the role of demonstrations: What makes in-context learning work? In: Proceedings of the 2022 Conference on Empirical Methods in Natural Language Processing. pp. 11048--11064 (2022)

\bibitem{owlvit}
Minderer, M., Gritsenko, A., Stone, A., Neumann, M., Weissenborn, D., Dosovitskiy, A., Mahendran, A., Arnab, A., Dehghani, M., Shen, Z., et~al.: Simple open-vocabulary object detection with vision transformers. arxiv 2022. arXiv preprint arXiv:2205.06230

\bibitem{slip}
Mu, N., Kirillov, A., Wagner, D., Xie, S.: Slip: Self-supervision meets language-image pre-training. In: European Conference on Computer Vision. pp. 529--544. Springer (2022)

\bibitem{clipit}
Narasimhan, M., Rohrbach, A., Darrell, T.: Clip-it! language-guided video summarization. Advances in Neural Information Processing Systems pp. 13988--14000 (2021)

\bibitem{flowers102}
Nilsback, M.E., Zisserman, A.: Automated flower classification over a large number of classes. In: Indian Conference on Computer Vision, Graphics \& Image Processing. pp. 722--729. IEEE (2008)

\bibitem{infonce}
Oord, A.v.d., Li, Y., Vinyals, O.: Representation learning with contrastive predictive coding. arXiv preprint arXiv:1807.03748  (2018)

\bibitem{orenes2021looking}
Orenes, I.: “looking at” negation: Faster processing for symbolic rather than iconic representations. Journal of Psycholinguistic Research (6),  1417--1436 (2021)

\bibitem{oxfordpets}
Parkhi, O.M., Vedaldi, A., Zisserman, A., Jawahar, C.: Cats and dogs. In: Proceedings of the IEEE/CVF Conference on Computer Vision and Pattern Recognition. pp. 3498--3505. IEEE (2012)

\bibitem{paszke2019pytorch}
Paszke, A., Gross, S., Massa, F., Lerer, A., Bradbury, J., Chanan, G., Killeen, T., Lin, Z., Gimelshein, N., Antiga, L., et~al.: Pytorch: An imperative style, high-performance deep learning library. Advances in Neural Information Processing Systems  (2019)

\bibitem{clip}
Radford, A., Kim, J.W., Hallacy, C., Ramesh, A., Goh, G., Agarwal, S., Sastry, G., Askell, A., Mishkin, P., Clark, J., et~al.: Learning transferable visual models from natural language supervision. In: International Conference on Machine Learning. pp. 8748--8763. PMLR (2021)

\bibitem{dalle2}
Ramesh, A., Dhariwal, P., Nichol, A., Chu, C., Chen, M.: Hierarchical text-conditional image generation with clip latents, 2022. arXiv preprint arXiv:2204.06125  (2022)

\bibitem{stablediff}
Rombach, R., Blattmann, A., Lorenz, D., Esser, P., Ommer, B.: High-resolution image synthesis with latent diffusion models. In: Proceedings of the IEEE/CVF Conference on Computer Vision and Pattern Recognition. pp. 10684--10695 (2022)

\bibitem{elephantintheroom}
Rosenfeld, A., Zemel, R.S., Tsotsos, J.K.: The elephant in the room. CoRR  (2018)

\bibitem{safavi2021negater}
Safavi, T., Zhu, J., Koutra, D.: Negater: Unsupervised discovery of negatives in commonsense knowledge bases. In: Proceedings of the 2021 Conference on Empirical Methods in Natural Language Processing. pp. 5633--5646 (2021)

\bibitem{schon2021negation}
Schon, C., Siebert, S., Stolzenburg, F.: Negation in cognitive reasoning. In: German Conference on Artificial Intelligence (K{\"u}nstliche Intelligenz). pp. 217--232. Springer (2021)

\bibitem{laion}
Schuhmann, C., Vencu, R., Beaumont, R., Kaczmarczyk, R., Mullis, C., Katta, A., Coombes, T., Jitsev, J., Komatsuzaki, A.: Laion-400m: Open dataset of clip-filtered 400 million image-text pairs. arXiv preprint arXiv:2111.02114  (2021)

\bibitem{flava}
Singh, A., Hu, R., Goswami, V., Couairon, G., Galuba, W., Rohrbach, M., Kiela, D.: {FLAVA:} {A} foundational language and vision alignment model. CoRR  (2021)

\bibitem{testoni2022artificial}
Testoni, A., Greco, C., Bernardi, R.: Artificial intelligence models do not ground negation, humans do. guesswhat?! dialogues as a case study. Frontiers in big Data p. 736709 (2022)

\bibitem{yfcc100m}
Thomee, B., Shamma, D.A., Friedland, G., Elizalde, B., Ni, K., Poland, D., Borth, D., Li, L.J.: Yfcc100m: The new data in multimedia research. Communications of the ACM (2),  64--73 (2016)

\bibitem{thrush2022winoground}
Thrush, T., Jiang, R., Bartolo, M., Singh, A., Williams, A., Kiela, D., Ross, C.: Winoground: Probing vision and language models for visio-linguistic compositionality. In: Proceedings of the IEEE/CVF Conference on Computer Vision and Pattern Recognition. pp. 5238--5248 (2022)

\bibitem{clipn}
Wang, H., Li, Y., Yao, H., Li, X.: Clipn for zero-shot ood detection: Teaching clip to say no. In: Proceedings of the IEEE/CVF International Conference on Computer Vision. pp. 1802--1812 (2023)

\bibitem{omnivl}
Wang, J., Chen, D., Wu, Z., Luo, C., Zhou, L., Zhao, Y., Xie, Y., Liu, C., Jiang, Y.G., Yuan, L.: Omnivl: One foundation model for image-language and video-language tasks. In: Koyejo, S., Mohamed, S., Agarwal, A., Belgrave, D., Cho, K., Oh, A. (eds.) Advances in Neural Information Processing Systems. vol.~35, pp. 5696--5710. Curran Associates, Inc. (2022)

\bibitem{xie2022an}
Xie, S.M., Raghunathan, A., Liang, P., Ma, T.: An explanation of in-context learning as implicit bayesian inference. In: International Conference on Learning Representations (2022)

\bibitem{negclip}
Yuksekgonul, M., Bianchi, F., Kalluri, P., Jurafsky, D., Zou, J.: When and why vision-language models behave like bags-of-words, and what to do about it? In: International Conference on Learning Representations (2022)

\bibitem{lit}
Zhai, X., Wang, X., Mustafa, B., Steiner, A., Keysers, D., Kolesnikov, A., Beyer, L.: Lit: Zero-shot transfer with locked-image text tuning. In: Proceedings of the IEEE/CVF Conference on Computer Vision and Pattern Recognition. pp. 18123--18133 (2022)

\end{thebibliography}
\newpage\appendix
\noindent{\Large{\textbf{Supplementary Material}}}\\

\noindent This supplementary material elaborates on our contributions and methodology and presents additional results. First, we provide details regarding our data generation process in Sec.~\ref{sec:data_gen}. Then, Sec.~\ref{sec:distractor_images} explains the methodology of obtaining distractor images for fine-tuning CLIP on our novel objectives. Lastly, Sec.~\ref{sec:more_results} provides additional results related to our experiments: negation understanding, zero-shot image classification, and general compositional understanding (replacing, adding, and swapping object, attributes, and relations).
\section{Prompting Scheme for CC-Neg Generation}
\label{sec:data_gen}
We use the image-labels subset of CC-3M as source for our image-caption pairs. Following this, the construction of CC-Neg is divided into two parts - negated caption generation and negative image mining.
\subsection{Generation of negated captions in CC-Neg}
\begin{figure*}[h!]
    \centering
    \includegraphics[width=\linewidth]{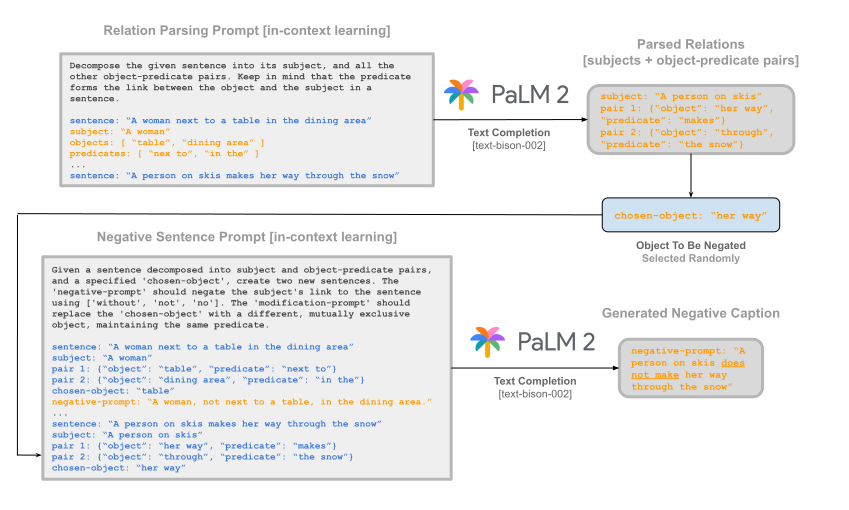}
    \caption{Illustration of using in-context learning with PaLM-2 to parse relations and subsequently generate negated captions. One meta-input-output demonstration is shown colored in blue and orange, each indicating the input and the output, respectively. Decomposition of positive captions into a subject and multiple predicate-object pairs is shown at the top. A random predicate-object pair is then selected and negated to generate a negative prompt, shown at the bottom.}
    \label{fig:relation_parsing}
\end{figure*}

The positive prompts from CC-3M are first decomposed using LLMs for relation parsing, and subsequently converted to negated captions, following the prompts and the process depicted in Fig.~\ref{fig:relation_parsing}. To summarize:
\begin{enumerate}
    \item{Positive image-caption pairs are extracted from the image-labels split of CC-3M.}
    \item{First, we prompt PaLM-2 to parse the relations inside positive prompts, i.e., captions related to the images. Each prompt is decomposed into a subject - who/what the sentence is about, as well as multiple predicate-object pair. Each predicate qualifies actions, or relates an object to the subject to specify a state of being.}
    \item{A single predicate-object pair is randomly selected to be negated for each sample.}
    \item{We then employ PaLM-2 to replace the selected predicate with one of \textit{no, not} and \textit{without} appropriately, and combine the subsequent atoms and relations into a negative caption. }
    \item{We drop samples with greater than 9 predicate-object pairs as this level of complexity is rarely found in real world data and is irrelevant from a human standpoint.}
\end{enumerate}
\section{Distractor Images for Finetuning}
\label{sec:distractor_images}
\begin{figure*}[h!]
    \centering
    \includegraphics[width=\linewidth]{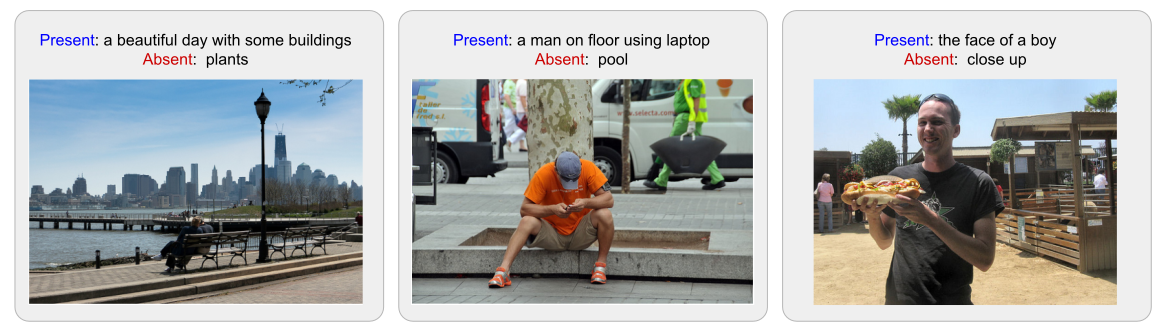}
    \caption{Examples of distractor images obtained from our process are shown above.}
    \label{fig:distractors}
\end{figure*}

This section describes our process of obtaining distractor images used in our fine-tuning process. 
Given a true caption $c$ and a negated caption $c'$ from CC-Neg, we first segregate concepts we know to be present from concepts that are absent in the scene, depicted in $c'$. Specifically, we take the subject $s$ and the negated object $o_n$ from the relation parsing output of PaLM-2 while generating $c'$. Next, given $s$ to be present and $o_n$ to be absent in the scene depicted by $c'$, we select a distractor image $I'$ from a large set of images (MSCOCO~\cite{coco}) $\boldsymbol{X}$ by
\begin{equation}
    I' = \argmax_{x \in \boldsymbol{X}} \quad \phi(x, s) - \phi(x, o_n)
\end{equation}
where $\phi(\cdot, \cdot)$ is the CLIP similarity function. Inspecting the results of this process, we obtain distractor images which represent the subject well but do not represent the negated object and its predicate. Examples of these distractor images are show in Fig.~\ref{fig:distractors}. While the negated captions can have more interpretations than their corresponding distractor images, we find distractor images to be suitable image negatives aimed at disentangling the existing effect of negations as well as other compositional deficiencies.
Notably, we do not consider this method of mining distractor images as a viable substitute for our framework CoN-CLIP. This is because (i) it requires an initial decomposition step to identify which semantics are present or absent, and (ii) it cannot be used to learn the effect of negations for a VLM. Consequently, this method sacrifices speed and acquisition of important new knowledge. Further, it cannot be used effectively for improving negation understanding in downstream applications of VLMs such as multimodal large language models (MLLMs) and text-to-image generation models (as mentioned in the main manuscript). This method is simply used to find suitable examples in the image modality which anchor the embeddings of negated captions for improved semantic disentanglement (with $\mathcal{L}_1$ and $\mathcal{L}_2$).
\section{Additional Results}
\label{sec:more_results}
This section presents additional results with an added baseline: LaCLIP~\cite{laclip}, a variant of CLIP which adds text augmentations during pretraining. This leads to a primary benefit to image classification by reducing overfitting to specific prompts~\cite{laclip}. Due to more exposure to language formats, we expect LaCLIP to show improved text understanding and compositionality, however, this is not an application proposed in~\cite{laclip}. Hence, to avoid confusion and maintain focus on the core contributions of CoN-CLIP, we omit this baseline from the main manuscript and provide the same here for comprehensiveness. We show LaCLIP's performance on negation understanding, general purpose compositionality, and zero-shot image classification. These results are shown alongside CLIP~\cite{clip}, NegCLIP~\cite{negclip}, BLIP~\cite{blip}, FLAVA~\cite{flava}, and all variants of CoN-CLIP, \textit{i.e.}, $\mathcal{L}_1, \mathcal{L}_2, \mathcal{L}_{12}$, and $\mathcal{L}_{conclip}$.
\subsection{Negation Understanding}
We present thorough evaluation of all VLMs on CC-Neg and its attributes: number of predicate-object pairs $\mathcal{K}$ and type of negation word used. These results are given in Table~\ref{tab:no_table}, Table~\ref{tab:not_table}, Table~\ref{tab:without_table}.
\begin{table}
    \centering
        \caption{Model Accuracies over CC-Neg evaluation subset samples that use "no" to specify negation. Scores are calculated over subset splits with the same number of predicate-object pairs, indicated by the number over each column. Split sizes are denoted inside parentheses.}
    \label{tab:no_table}
    \begin{tabular*}{\linewidth}{@{\extracolsep{\fill}}lccccc}
        \toprule
        \multirow{2}{*}{\textbf{Model Configuration}} & \multicolumn{5}{c}{\textbf{No. of predicate-object pairs}}\vspace{4pt}\\
        \cmidrule{2-6}
         & \#1 (166) & \#2 (160) & \#3 (75) & \#4 (28) & \#5 (8)\\
        \midrule
        CLIP & 70.48 & 68.13 & 69.33 & 53.57 & 37.5 \\
        LaCLIP & 69.88 & 81.25 & 64.0 & 67.86 & 50.0 \\
        BLIP & 76.51 & 63.12 & 72.0 & 42.86 & 50.0 \\
        FLAVA & 62.05 & 58.75 & 57.33 & 57.14 & 62.5 \\
        NegCLIP & 69.28 & 64.38 & 64.0 & 57.14 & 25.0 \\
        CoN-CLIP & 96.99 & 93.13 & 92.0 & 82.14 & 75.0 \\
        \bottomrule
    \end{tabular*}
\end{table}
\begin{table}
    \centering
        \caption{Model Accuracies over CC-Neg evaluation subset samples that use "not" to specify negation. Scores are calculated over subset splits with the same number of predicate-object pairs, indicated by the number over each column. Split sizes are denoted inside parentheses.}
    \label{tab:not_table}
    \begin{tabular*}{\linewidth}{@{\extracolsep{\fill}}lccccc}
        \toprule
        \multirow{2}{*}{\textbf{Model Configuration}} & \multicolumn{5}{c}{\textbf{No. of predicate-object pairs}}\vspace{4pt}\\
        \cmidrule{2-6}
         & \#1 (11910) & \#2 (10081) & \#3 (3047) & \#4 (722) & \#5 (188)\\
        \midrule
        CLIP & 68.13 & 64.94 & 59.11 & 59.83 & 56.91\\
        LaCLIP & 71.08 & 66.52 & 58.62 & 54.02 & 53.19\\
        BLIP & 60.76 & 60.07 & 54.25 & 56.23 & 50.00\\
        FLAVA & 59.73 & 54.58 & 50.51 & 50.42 & 52.13\\
        Neg-CLIP & 66.32 & 63.21 & 57.66 & 55.12 & 55.32\\
        CoN-CLIP & 99.88 & 99.84 & 99.64 & 99.03 & 99.47\\
        \bottomrule
    \end{tabular*}
\end{table}
\begin{table}
    \centering
        \caption{Model Accuracies over CC-Neg evaluation subset samples that use "without" to specify negation. Scores are calculated over subset splits with the same number of predicate-object pairs, indicated by the number over each column. Split sizes are denoted inside parentheses.}
    \label{tab:without_table}
    \begin{tabular*}{\linewidth}{@{\extracolsep{\fill}}lccccc}
        \toprule
        \multirow{2}{*}{\textbf{Model Configuration}} & \multicolumn{5}{c}{\textbf{No. of predicate-object pairs}}\vspace{4pt}\\
        \cmidrule{2-6}
         & \#1 (4850) & \#2 (5466) & \#3 (2328) & \#4 (680) & \#5 (203)\\
        \midrule
        CLIP & 66.14 & 66.39 & 65.38 & 63.09 & 70.94 \\
        LaCLIP & 69.36 & 59.79 & 56.36 & 55.44 & 48.28 \\
        BLIP & 70.12 & 67.54 & 64.18 & 62.06 & 67.98 \\
        FLAVA & 65.98 & 63.57 & 61.77 & 61.62 & 58.13 \\
        Neg-CLIP & 62.12 & 60.92 & 57.39 & 56.32 & 60.59 \\
        CoN-CLIP & 99.88 & 99.82 & 99.91 & 99.71 & 100.0 \\
        \bottomrule
    \end{tabular*}
\end{table}
\newpage\subsection{Con-CLIP Ablation Study on CC-Neg}
\begin{table}
    \centering
        \caption{CoN-CLIP ablation study over CC-Neg evaluation subset samples that use "no" to specify negation. Scores are calculated over subset splits with the same number of predicate-object pairs, indicated by the number over each column. Split sizes are denoted inside parentheses.}
    \label{tab:no_table_ablation}
    \begin{tabular*}{\linewidth}{@{\extracolsep{\fill}}lccccc}
        \toprule
        \multirow{2}{*}{\textbf{Model Configuration}} & \multicolumn{5}{c}{\textbf{No. of predicate-object pairs}}\vspace{4pt}\\
        \cmidrule{2-6}
         & \#1 (166) & \#2 (160) & \#3 (75) & \#4 (28) & \#5 (8)\\
        \midrule
        CoNCLIP ViT-B/32 $\mathcal{L}_{1}$ & 96.99 & 93.13 & 90.67 & 85.71 & 62.5 \\
        CoNCLIP ViT-B/32 $\mathcal{L}_{2}$ & 65.06 & 58.75 & 57.33 & 32.14 & 25.0 \\
        CoNCLIP ViT-B/32 $\mathcal{L}_{12}$ & 95.18 & 93.13 & 90.67 & 78.57 & 75.0 \\
        CoN-CLIP ViT-B/32 $\mathcal{L}_{conclip}$ & 96.99 & 93.13 & 92.0 & 82.14 & 75.0 \\
        \bottomrule
    \end{tabular*}
\end{table}
\begin{table}
    \centering
        \caption{CoN-CLIP ablation study over CC-Neg evaluation subset samples that use "not" to specify negation. Scores are calculated over subset splits with the same number of predicate-object pairs, indicated by the number over each column. Split sizes are denoted inside parentheses.}
    \label{tab:not_table_ablation}
    \begin{tabular*}{\linewidth}{@{\extracolsep{\fill}}lccccc}
        \toprule
        \multirow{2}{*}{\textbf{Model Configuration}} & \multicolumn{5}{c}{\textbf{No. of predicate-object pairs}}\vspace{4pt}\\
        \cmidrule{2-6}
         & \#1 (11910) & \#2 (10081) & \#3 (3047) & \#4 (722) & \#5 (188)\\
        \midrule
        CoNCLIP ViT-B/32 $\mathcal{L}_{1}$ & 99.84 & 99.84 & 99.74 & 99.58 & 99.47 \\
        CoNCLIP ViT-B/32 $\mathcal{L}_{2}$ & 55.89 & 52.16 & 49.56 & 48.20 & 46.81 \\
        CoNCLIP ViT-B/32 $\mathcal{L}_{12}$ & 99.83 & 99.81 & 99.67 & 99.17 & 98.94 \\
        CoN-CLIP ViT-B/32 $\mathcal{L}_{conclip}$ & 99.88 & 99.84 & 99.64 & 99.03 & 99.47\\
        \bottomrule
    \end{tabular*}
    \vspace{-20pt}
\end{table}
\begin{table}[h!]
    \centering
        \caption{CoN-CLIP ablation study over CC-Neg evaluation subset samples that use "without" to specify negation. Scores are calculated over subset splits with the same number of predicate-object pairs, indicated by the number over each column. Split sizes are denoted inside parentheses.}
    \label{tab:without_table_ablation}
    \begin{tabular*}{\linewidth}{@{\extracolsep{\fill}}lccccc}
        \toprule
        \multirow{2}{*}{\textbf{Model Configuration}} & \multicolumn{5}{c}{\textbf{No. of predicate-object pairs}}\vspace{4pt}\\
        \cmidrule{2-6}
         & \#1 (4850) & \#2 (5466) & \#3 (2328) & \#4 (680) & \#5 (203)\\
        \midrule
        CoNCLIP ViT-B/32 $\mathcal{L}_{1}$ & 99.94 & 99.89 & 99.79 & 99.85 & 100.0 \\
        CoNCLIP ViT-B/32 $\mathcal{L}_{2}$ & 62.93 & 61.23 & 58.76 & 55.00 & 60.59 \\
        CoNCLIP ViT-B/32 $\mathcal{L}_{12}$ & 99.92 & 99.87 & 99.74 & 99.85 & 100.0 \\
        CoN-CLIP ViT-B/32 $\mathcal{L}_{conclip}$ & 99.88 & 99.82 & 99.91 & 99.71 & 100.0 \\
        \bottomrule
    \end{tabular*}
\end{table}
\subsection{General Purpose Compositional Understanding}
\begin{table}[h!]
    \caption{Evaluating CoN-CLIP on SugarCREPE alongside CLIP on R@1. Highest performance for a fold and CLIP backbone are \underline{underlined} and \textit{italicised} respectively.}
    \label{tab:sugarcrepe}
    \centering
    \begin{tabular*}{0.9\linewidth}{@{\extracolsep{\fill}}lccccccc}
       \toprule
       \multirow{2}{*}{\textbf{Model}}  &  \multicolumn{3}{c}{\textbf{Replace}} & \multicolumn{2}{c}{\textbf{Add}} & \multicolumn{2}{c}{\textbf{Swap}}\\
        \cmidrule{2-4}\cmidrule{5-6}\cmidrule{7-8}
        & {Object} & {Attribute} & {Relation} & {Object} & {Attribute} & {Object} & {Attribute}\\
        \midrule
        \textbf{CLIP} & & & & & & & \\
        ViT-B/16 & 93.28 & 80.83 & \textit{66.00} & 78.32 & 66.61 & \textit{59.59} & 64.41\\
        ViT-B/32 & 90.79 & 80.07 & \textit{68.99} & 76.91 & 68.35 & 60.81 & 63.06\\
        ViT-L/14 & 94.06 & 79.18 & 65.07 & 78.17 & 71.38 & 60.00 & 62.16\\
        \midrule
        \textbf{LaCLIP} & & & & & & & \\
        ViT-B/16 & 93.22 & 79.69 & 58.32 & 77.44 & 66.18 & 59.59  &  59.15\\
        ViT-B/32 & 91.28 & 77.66 & 57.75 & 75.41 & 64.01 & 55.51 & 59.55 \\
        ViT-L/14 & 93.28 & 81.09 & 61.73 & 81.57 & 73.55 & 62.44 & 58.70 \\
        \midrule
        \textbf{CoN-CLIP} & & & & & & & \\
        ViT-B/16 & \textit{93.58} & \textit{80.96} & 63.30 & \textit{87.29} & \textit{79.62} & 59.18 & \textit{65.16} \\
        ViT-B/32 & \textit{91.76} & \textit{80.96} & 66.28 & \textit{87.92} & \textit{78.03} & \textit{63.67} & \textit{66.96}\\
        ViT-L/14 & \textit{\underline{95.31}} & \textit{\underline{81.72}} & \textit{\underline{66.99}} & \textit{\underline{90.15}} & \textit{\underline{77.60}} & \textit{\underline{65.36}} & \textit{\underline{63.06}} \\
    \bottomrule
    \\
    \end{tabular*}
    \vspace{-15pt}
\end{table}
\subsection{CoN-CLIP Ablation Study on SugarCREPE}
\begin{table}[h!]
    \centering
        \caption{Evaluating CoN-CLIP ablations on SugarCREPE.}
    \label{tab:sugarcrepe_conclip_ablations}
    \begin{tabular*}{0.9\linewidth}{@{\extracolsep{\fill}}lccccccc}
       \toprule
       \multirow{2}{*}{\textbf{Model}}  &  \multicolumn{3}{c}{\textbf{Replace}} & \multicolumn{2}{c}{\textbf{Add}} & \multicolumn{2}{c}{\textbf{Swap}}\\
        \cmidrule{2-4}\cmidrule{5-6}\cmidrule{7-8}
        & {Object} & {Attribute} & {Relation} & {Object} & {Attribute} & {Object} & {Attribute}\\
        \midrule
        \textbf{CoN-CLIP} & & & & & & & \\        
        $\mathcal{L}_{1}$ & 91.71 & 80.58 & 65.79 & 84.82 & 79.62 & 59.18 & 64.11 \\
        $\mathcal{L}_{2}$ & 91.65 & 81.22 & 65.29 & 88.02 & 82.51 & 64.49 & 67.27 \\
        $\mathcal{L}_{12}$  & 91.46 & 81.85 & 68.35 & 86.81 & 79.77 & 61.63 & 66.52 \\
        $\mathcal{L}_{conclip}$ & 91.76 & 80.96 & 66.28 & 87.92 & 78.03 & 63.67 & 66.96\\
    \bottomrule
    \\
    \end{tabular*}
\end{table}
\newpage\subsection{Zero-shot Image Classification}
\begin{table*}[h!]
    \caption{Evaluation of CoN-CLIP ($\mathcal{L}_{conclip}$) on zero-shot image classification shows improvements across all datasets. Here, highest accuracy values for a dataset are \underline{underlined}, while highest accuracy values for a CLIP backbone are given in \textit{italics}.}
    \label{tab:zeroshot_table}
    \centering
    \begin{tabular*}{\linewidth}{@{\extracolsep{\fill}}lcccccccc}
        \toprule
        \cmidrule{2-9}
        \multirow{2}{*}{\textbf{Model}} & \textbf{ImageNet} & \textbf{Caltech} & \textbf{Flowers} & \textbf{CIFAR} & \textbf{Food} & \textbf{Stanford} & \textbf{Oxford} & \textbf{CIFAR} \\
         & \textbf{1k} & \textbf{101} & \textbf{102} & \textbf{100} & \textbf{101} & \textbf{Cars} & \textbf{Pets} & \textbf{10} \\
        \midrule 
        \small{\textbf{CLIP}} & & & & & & & & \\
        \small{ViT-B/16} & \small{68.35} & \small{82.56} & \small{64.14} & \small{53.54} & \small{86.89} & \small{61.68} & \small{81.82} & \small{88.23}\\
        \small{ViT-B/32} & \small{63.36} & \small{81.50} & \small{60.50} & \small{55.18} & \small{81.15} & \small{58.33} & \small{80.08} & \small{88.97}\\
        \small{ViT-L/14} & \small{75.51} & \small{81.80} & \small{72.42} & \small{65.95} & \small{92.10} & \small{74.64} & \small{88.06} & \small{\underline{\textit{91.40}}}\\
        
        \midrule 
        \small{\textbf{LaCLIP}} & & & & & & & & \\
        \small{ViT-B/16} & \small{67.20} & \small{87.39} & \small{66.11} & \textit{67.82} & \small{82.82} & \textit{85.61} & \small{83.95} & \small{91.82}\\ 
        \small{ViT-B/32} & \small{62.01} & \textit{87.95} & \small{62.76} & \textit{65.56} & \small{75.61} & \textit{80.01} & \small{80.84} & \small{\underline{\textit{91.38}}}\\ 
        \small{ViT-L/14} & \small{74.50} & \small{\underline{\textit{89.81}}} & \small{\underline{\textit{75.85}}} & \small{\underline{\textit{79.32}}} & \small{90.28} & \small{\underline{\textit{90.77}}} & \small{89.29} & \small{\underline{\textit{96.69}}}\\ 
        
        \midrule  
        \small{\textbf{CoN-CLIP}} & & & & & & & & \\
        \small{ViT-B/16} & \textit{68.95} & \textit{87.62} & \textit{66.69} & \small{64.49} & \textit{88.13} & \small{62.08} & \textit{85.45} & \small{90.88}\\ 
        \small{ViT-B/32} & \textit{63.36} & \small{86.91} & \textit{64.74} & \small{62.31} & \textit{83.39} & \small{58.84} & \textit{81.66} & \small{90.45}\\
        \small{ViT-L/14} & \small{\underline{\textit{75.93}}} & \small{87.90} & \small{75.12} & \small{75.39} & \small{\underline{\textit{93.01}}} & \small{76.17} & \small{\underline{\textit{89.32}}} & \small{95.05}\\
        \bottomrule
        \\
    \end{tabular*}
    \vspace{-15pt}
\end{table*}
\begin{table}[h!]
    \centering
\caption{Comparing $\Delta$ values averages across $8$ image classification datasets (refer Sec.~\ref{sec:negations} of the main manuscript) using the ViT-B/32 backbone.}
    \label{tab:my_label}
    \begin{tabular}{lc}
        \toprule
        \textbf{Model} & \textbf{Mean} $\Delta$ (\%) $\uparrow$ \\
        \midrule
        CLIP & 0.98 \\
        {LaCLIP} & -0.99\\
        NegCLIP & -1.01 \\
        \underline{CoN-CLIP} & \underline{62.03} \\
        \bottomrule
        \\
    \end{tabular}
\end{table}
\end{document}